\documentclass[conference,a4paper]{IEEEtran}

\usepackage[utf8]{inputenc}

\include{basic-includes}
\usepackage{hyperref}
\usepackage{graphicx}
\usepackage{caption, subcaption}
\usepackage{tabularx, booktabs}

\usepackage{stfloats}
\usepackage{multicol}

\usepackage{color}

\newcommand{\covid}{COVID-19}

\usepackage{cite}
\usepackage[section]{placeins}

\newcommand\blfootnote[1]{%
  \begingroup
  \renewcommand\thefootnote{}\footnote{#1}%
  \addtocounter{footnote}{-1}%
  \endgroup
}

\def\mywidth{.95\columnwidth}

\begin{document}

\title{Hands Off: A Handshake Interaction Detection and Localization Model for COVID-19 Threat Control }

\author{\IEEEauthorblockN{\\ \\ \\ \\ \\ \\ \\}}
\author{\IEEEauthorblockN{
A. S. Jameel Hassan$^{\dagger}$, 
Suren Sritharan$^{\ddagger}$, 
Gihan Jayatilaka$^{\dagger}$,\\ Roshan I. Godaliyadda$^{\dagger}$,
Parakrama B. Ekanayake$^{\dagger}$,
Vijitha Herath$^{\dagger}$, 
Janaka B. Ekanayake$^{\dagger}$\\}
\IEEEauthorblockA{
\emph{$^{\dagger}$Department of Electrical and Electronic Engineering, University of Peradeniya, Sri Lanka}\\ 
\emph{$^{\ddagger}$School of Computing and IT, Sri Lanka Technological Campus, Sri Lanka}\\
}
\IEEEauthorblockA{ \tt{\{jameel.hassan.2014, suren.sri, gihanjayatilaka\}@eng.pdn.ac.lk},\\
\tt{\{roshangodd, mpb.ekanayake\}@ee.pdn.ac.lk, \{vijitha, ekanayakej\}@eng.pdn.ac.lk,}
}
}

\maketitle

\begin{abstract}
\blfootnote{This paper has been accepted for ICIIS 2021}
The \covid\ outbreak has affected millions of people across the globe and is continuing to spread at a drastic scale.
Out of the numerous steps taken to control the spread of the virus, social distancing has been a crucial and effective practice. 
However, recent reports of social distancing violations suggest the need for non-intrusive detection techniques to ensure safety in public spaces.
In this paper, a real-time detection model is proposed to identify handshake interactions in a range of realistic scenarios with multiple people in the scene and also detect multiple interactions in a single frame.
The efficacy of the proposed model was evaluated across two different datasets on more than 3200 frames, thus enabling a robust localization model in different environments. The proposed model is the first dyadic interaction localizer in a multi-person setting, which enables it to be used in public spaces to identify handshake interactions and thereby identify and mitigate \covid\ transmission.

\begin{IEEEkeywords}
\covid, deep learning, human-human interactions, dyadic interaction localization
\end{IEEEkeywords}

\end{abstract}

\section{Introduction}

The novel \covid\ virus is one of the biggest threat to global health since the Spanish flu in 1918. As of July 2021 nearly 196 million people have been infected and more than 4 million people have succumbed to death due to the virus \cite{worldometer}. Vaccination has been identified as the most effective measure by the World Health Organization (WHO) to curtail the spread of the virus \cite{WHOvaccine}. However, complete vaccination of the entire global population has not been possible due to varying production and logistic issues. Therefore, the key measure taken for the curtailment of the spread of \covid\ has been social distancing. \par

Social distancing has been found to be a promising approach towards mitigating the virus spread \cite{SDcovid}. Nevertheless, humans as a social species, tend to deviate from such constrained behavior \cite{SDmental} for prolonged periods of time. Thus, it is crucial to identify such breach of social distancing protocols in order to ensure the safety of the society. Importantly, human-human interactions need to be ensured minimal as it is the most severe form of breach which are also the easiest to avoid. Moreover, a simple greeting is the often the initial breach of social distancing. Therefore, identifying and localizing such interactions such as from a CCTV footage will enable to create a framework to prevent such breach of social distancing measures. \par

Identification of human interactions often referred to as dyadic interactions (interactions between two people) has been explored in the action recognition domain. Action recognition has moved from an object detection/tracking problem \cite{samithaH,rupasingheA} in to a multi-class classification problem. Dyadic interaction detection has spawned from human action recognition in computer vision literature. Most of action recognition has focused on a single person in the frame performing a specific action such as running, walking, jumping etc \cite{actionrecog}. Recently works have focused on behavior/activity recognition of multiple people in the frame, such as in a game \cite{FeiFei}. \par

The use of limb positions to identify interactions was presented in \cite{bourdev}. The idea stemmed from the concept that each interaction presented unique limb positioning. As a next step, considering the gross body movement and proximity measures was done in \cite{perez}. This is done in a multi-step manner where the person localization is used for the interaction identification. This poses a drawback in interaction localization as the error in the first stage of person localization can extend to the next stage. This has been improved by considering this a multiple instance learning problem by \cite{ikizler} since not all frames in an interaction are considered informative. \par

Bag of visual words method has also been used to identify body movements. Local features from this are pooled and a mapping is generated from this to interactions in \cite{STIP}. Part-based models such as deformable part model (DPM) \cite{DPM} has been proven extremely effective in people and body part detection and localization prior to neural networks. The use of interaction specific DPMs to identify people in specific poses is done in \cite{Van-Gemeren1}. An extension of this work using spatio-temporal DPMs to localize dyadic interactions has been presented in \cite{Van-Gemeren2}. This has been one of the few works that localizes the interaction itself instead of the actors. \par

\begin{figure*}[b]
    \centering
    \includegraphics[width=\textwidth]{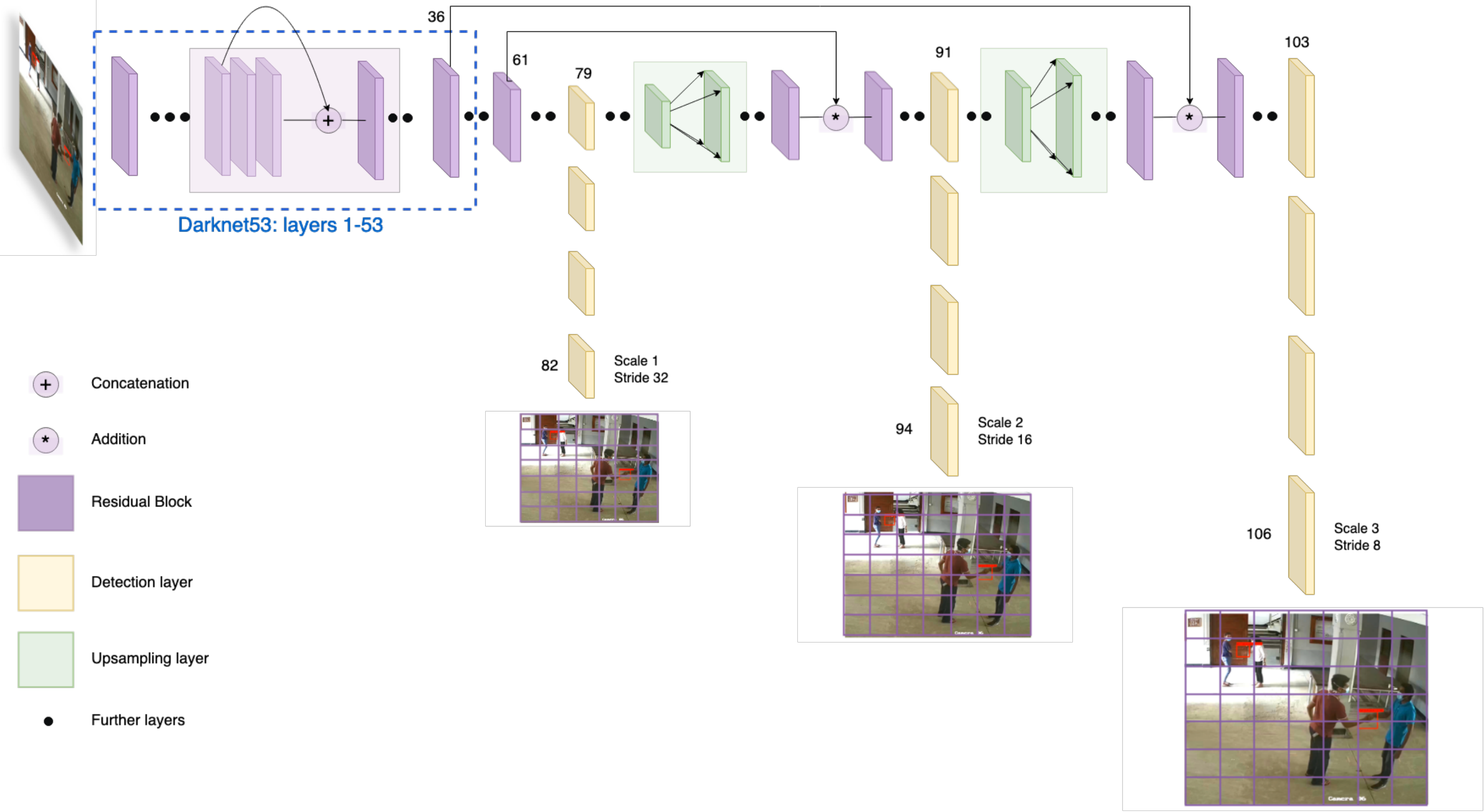}
    \caption{YOLOv3 architecture.}
    \label{fig:yolov3}
\end{figure*}

The advent of neural networks has drastically overtaken DPM techniques in detection problems. The YOLO network \cite{YOLO} is a highly robust neural network capable of detecting 80 classes in real-time (78 FPS). In \cite{YOLOcomproc} a human activity recognition model has been formulated using the YOLO network on the LIRIS dataset \cite{LIRIS}. Most notably, this model can perform the localization in real-time which is crucial depending on the need. A recurrent neural network (RNN) based spatio-temporal attention mechanism for human interaction recognition is performed in \cite{baradel}. This model incorporates attention to the hands of the body to identify the interaction. However, one of the main drawbacks of this and other methods is the absence of real-time detection. The above cited works except \cite{Van-Gemeren1,Van-Gemeren2} in dyadic interaction detection consider a video feed/frame and classify it to the given class of interaction or identify the actors where the localization of the interaction is not considered. This localization too is performed only with two persons in the frame. \par 

The major contributions of this paper are as follows. In this paper, the first human interaction localization model in a multi-person setting is proposed. A handshake interaction localization model in real-time to mitigate the threat for \covid\ is presented using computer vision in a non-intrusive technique. This ensures a scalable, robust model that can be used in public spaces and work environments to mitigate the spread of \covid.

\section{Proposed solution}

A convolutional neural network (CNN) based model is proposed to identify and localize handshake interactions in a multi-person setting for wall mounted CCTV video footage. The model architecture used is the YOLO network with training and testing performed using a novel dataset and the UT-interaction (UTI) \cite{UT-Interaction-Data,UT2} dataset. \par

\subsection{YOLO network}
The YOLO network is a state-of-the-art (SOTA) CNN in object detection. It was the pioneering work in creating a one-stage detection network for the object detection task. The key change in YOLOv3 \cite{yolov3} was the approach to divide the image into grids (such as 13$\times$13), and then predict a fixed number (such as 3) of bounding boxes for each grid cell. The bounding box is predicted with the relevant class and object confidence score. The architecture of the YOLOv3 network is shown in Fig.\ref{fig:yolov3}. \par

The YOLOv3 architecture makes prediction at three stages in the neural network depth as seen in Fig.\ref{fig:yolov3}. This enables detection of objects of all sizes, which was the main drawback in previous versions. The first stage detector outputting a $13\times13$ grid is better at predicting larger images, while the $26\times26$ grid predicts medium sized images and the $52\times52$ grid prediction in stage three is best at predicting small images. The image input (resized to $416\times416$ passes through the convolutional layers to output a tensor of shape $h \times h\times18$. Here $h$ is the number of grid cells along one axis and 18 corresponds to $3\times(5+1)$, where $3$ is the number of bounding boxes predicted in one grid cell, $5$ is the number of bounding box attributes and $1$ is the number of classes. The bounding box attributes are the coordinates of the four vertices and the objectness score.\par

\begin{figure*}[t]
    \centering
    \begin{subfigure}[b]{0.32\textwidth}
        \includegraphics[width=\textwidth]{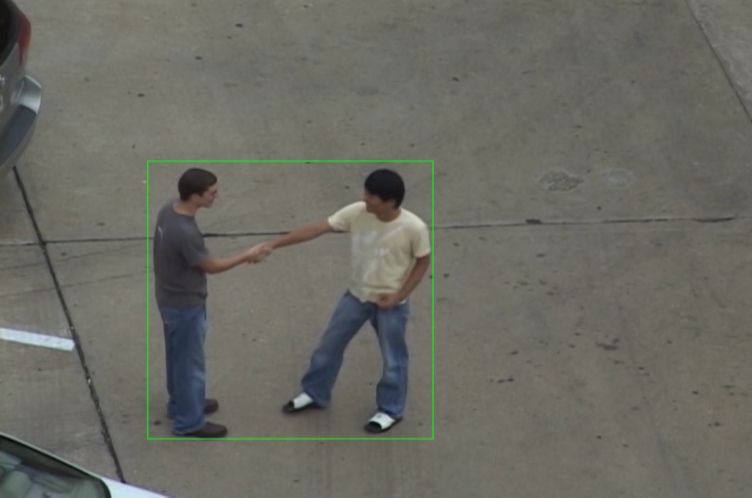}
        \caption{Original ground truth for UTI.}
    \end{subfigure}
    \hfill
    \begin{subfigure}[b]{0.32\textwidth}
        \includegraphics[width=\textwidth]{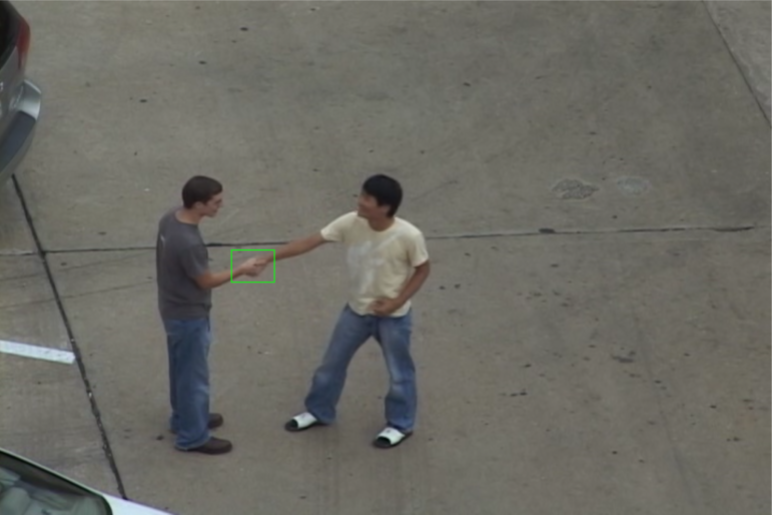}
        \caption{Created ground truth for UTI.}
    \end{subfigure}
    \hfill
    \begin{subfigure}[b]{0.33\textwidth}
        \includegraphics[width=\textwidth]{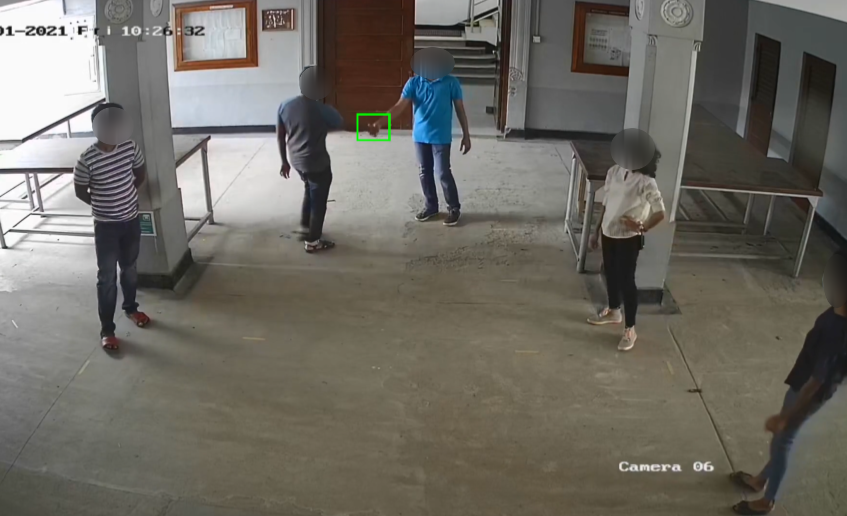}
        \caption{Ground truth of Shakes.}
        \label{fig:deee}
    \end{subfigure}
    \caption{Dataset ground truth annotations.}
    \label{fig:uti-before-after}
\end{figure*}

\subsection{Model Training}

In order to train the YOLO network for handshake interaction detections, a suitable dataset is required. There are few datasets in the action recognition domain for computer vision. However, the handshake interactions in these datasets are minimal and even then, the ground truth for such datasets are not for the localization problem but for actors identification. Furthermore, existing datasets for dyadic interactions have only two people in the frame. Since our motivation is to identify interactions to combat COVID-19, a video footage with multiple people in the frame, where dyadic interactions occur is necessary. Therefore, a dataset rich in context to tackle the problem of human interaction identification in a multi-person setting was created. The existing UTI dataset was also used in the framework by relabelling the handshake interactions for the localization problem. \par

\subsection{Datasets}
The existing UTI dataset was re-labelled by marking the interactions in each frame. Fig.\ref{fig:uti-before-after} shows the original ground truth and the created ground truth data for the UTI dataset. 

Due to the scarcity of handshake interactions, a new dataset was created in the university premises using wall mounted CCTVs. This consisted of 10 videos each spanning nearly 1500 frames. We refer to this dataset as the "Shakes dataset". This consists a multi-person setting and also multiple interactions in the same instance in many frames. A sample frame is shown in Fig.\ref{fig:deee}. \par

\subsection{Training Using Transfer Learning}
In order to train the YOLO network for the handshake interactions, the darknet53 (highlighted by a cyan dotted rectangle) Fig.\ref{fig:yolov3}, referred to as the YOLO backbone was initialized with weights obtained by training on the Imagenet dataset \cite{imagenet}. Then, 3000 images of hands from the open images database \cite{openimages} were used for training the YOLO network as the first stage, since a larger distribution of images was available here. Using the weights of the network from this training phase, the handshake images from the Shakes and UTI datasets were trained. The transfer learning approach was used as the handshake interactions were from a smaller distribution. Out of 20 videos, 17 videos from the UTI dataset and 5 out of the 10 videos from the Shakes dataset were used. While the number of videos from the UTI dataset is higher, the number of frames were maintained approximately equal. 

\section{Results and Discussion}

The model was evaluated using both the aforementioned datasets. The Average precision (AP) and the Mean average precision was used as the evaluation metric, as prominent object detection competitions such as PASCAL VOC challenge \cite{pascalvoc}, COCO detection challenge \cite{COCO} and the Google Open Images dataset \cite{openimages} competition use these metrics as key parameters in evaluating the detector performance. \par

The Average precision (AP) is the precision value averaged across varying recall values between 0 and 1. This is computed using area under the curve (AUC) of the precision vs recall curve, plotted as a function of the confidence threshold of detection with a constant intersection over union (IoU) for the bounding box threshold \cite{padilla2}. This IoU threshold is usually maintained at 0.5 in object detection tasks.

\begin{figure}[t]
    \centering
    \captionsetup[subfigure]{aboveskip=2pt, belowskip=2pt}
    
    \begin{subfigure}{0.48\textwidth}
        \centering
        \includegraphics[width=\textwidth]{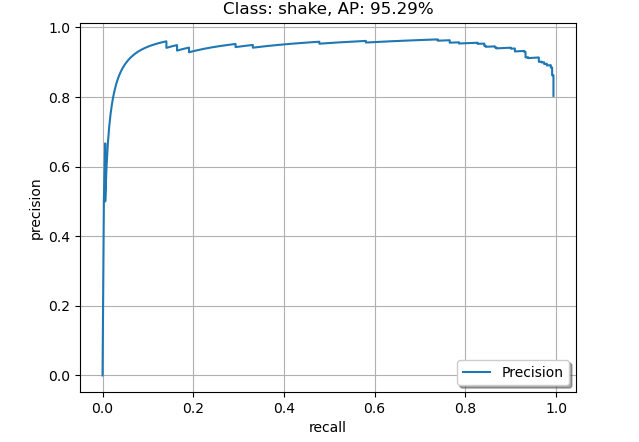}
        \caption{UTI dataset}
    \end{subfigure}
    \hfill
    \begin{subfigure}{0.48\textwidth}
        \centering
        \includegraphics[width=\textwidth]{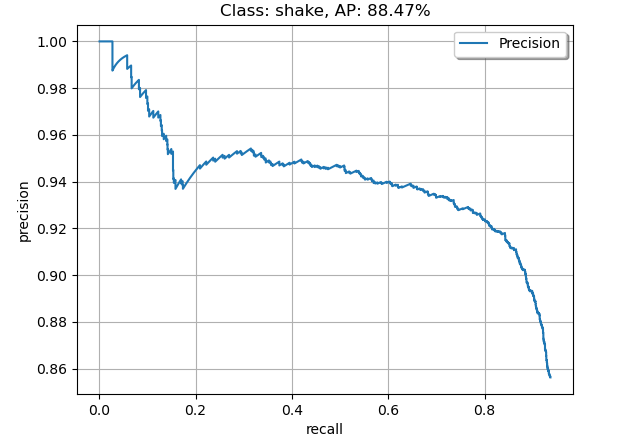}
        \caption{Shakes dataset}
    \end{subfigure}
    \caption{Precision vs Recall curves for handshake localizer for UTI and Shakes dataset.}
    \label{fig:PvsR_shake}
\end{figure}

The performance of the model in detecting handshake interactions was evaluated on the UTI and Shakes dataset separately and is tabulated in Table \ref{tab:handshake}. 3 videos containing 418 frames from the UTI dataset and 5 videos with 2786 frames from the Shakes dataset were considered for this purpose. The AP value for the UTI dataset was 95.29\% and for the Shakes dataset was 88.47\%. The precision vs recall curves for the UTI dataset and the Shakes dataset are shown in Fig.\ref{fig:PvsR_shake}. 

\begin{table}[tbh]
    \caption{Performance metrics of handshake detection}
    \begin{center}
    \begin{tabular}[\columnwidth]{m{0.5\columnwidth} m{0.3\columnwidth}}
        \toprule
        \textbf{Dataset} & \textbf{AP}/\% \\
        \bottomrule
        UT-interaction & 95.29\\
        \midrule
        Shakes & 88.47\\
        \bottomrule
    \end{tabular}
    \end{center}
    \label{tab:handshake}
\end{table}

\begin{figure}[t]
    \centering
    \captionsetup[subfigure]{aboveskip=2pt, belowskip=2pt}
    
    \begin{subfigure}[t]{\columnwidth}
        \includegraphics[width=\textwidth]{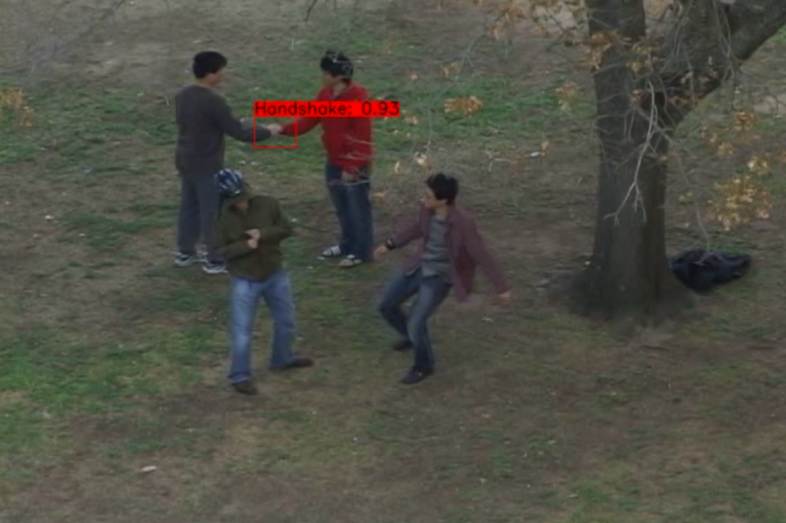}
        \caption{UTI dataset frame}
        \label{fig:uti-shake1}
    \end{subfigure}%
    
    \begin{subfigure}[t]{\columnwidth}
        \includegraphics[width=\textwidth]{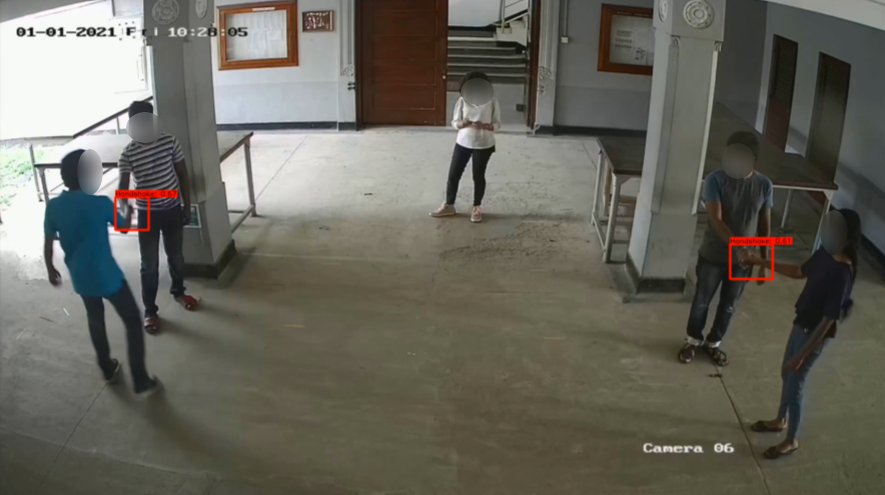}
        \caption{Shakes dataset - two handshakes.}
        \label{fig:deee-2shakes}
    \end{subfigure}%
    
    \begin{subfigure}[t]{\columnwidth}
        \includegraphics[width=\textwidth]{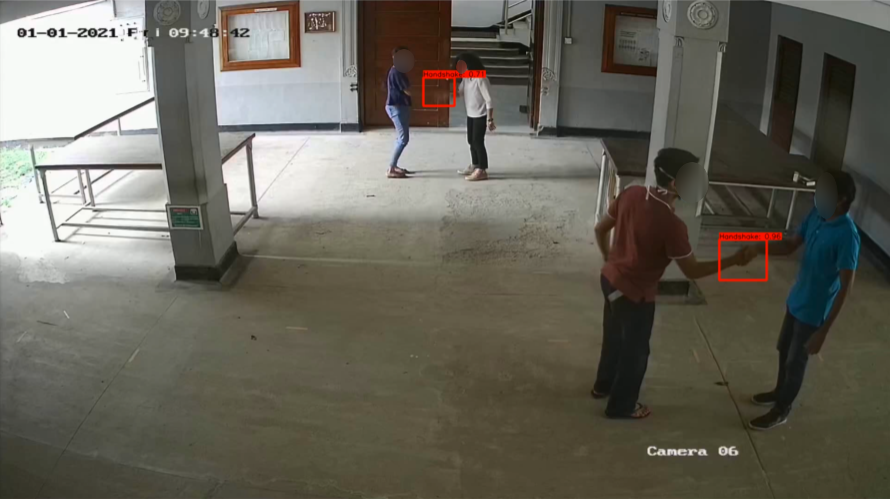}
        \caption{Handshakes at different distances from the camera.}
        \label{fig:deee-2shakes-scale}
    \end{subfigure}%
    
    \caption{Handshake interaction detection localizations.}
    \label{fig:res-shakes}
\end{figure}

\begin{figure*}[htb!]
    \centering
     \captionsetup[subfigure]{aboveskip=2pt, belowskip=2pt}
     
    \begin{subfigure}[t]{\columnwidth}
        \centering
        \includegraphics[width=\mywidth]{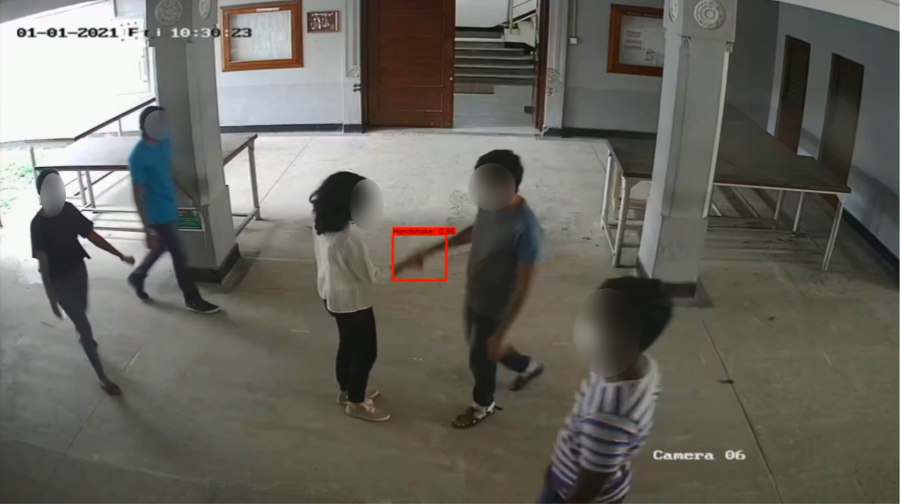}
        \caption{Handshake in corridor.}
    \end{subfigure}%
    \hfill
    \begin{subfigure}[t]{\columnwidth}
        \centering
        \includegraphics[width=\mywidth]{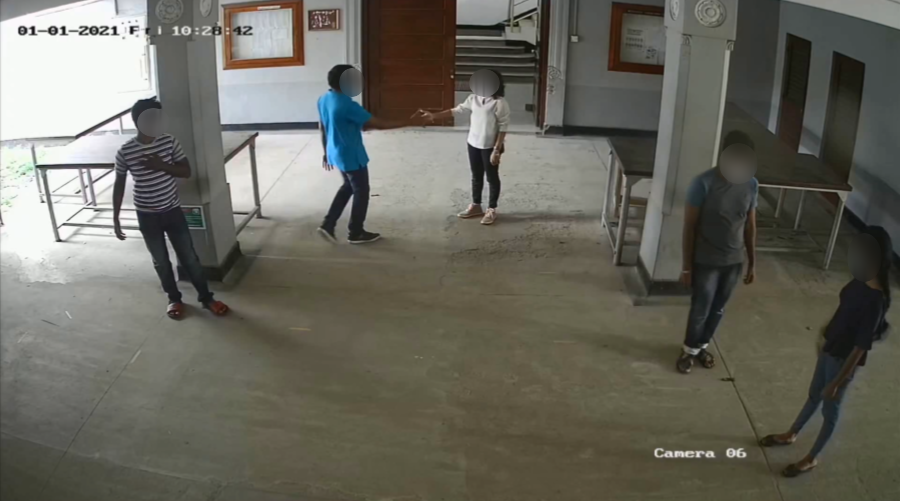}
        \caption{Fake handshake by occlusion.}
        \label{fig:fakeshake}
    \end{subfigure}

    \caption{Detection localizations in busy settings and occlusion cases.}
    \label{fig:res-rand-shakes}
\end{figure*}

Few frames of detection and localization of handshake interactions from the UTI and the Shakes dataset are shown in Fig.\ref{fig:res-shakes}. It can be observed that the neural network can identify more than just a single handshake interaction in the frame. It is also able to identify interactions even at different scales as seen in Fig.\ref{fig:deee-2shakes-scale}. A more realistic setting such as interactions in a busy public space is shown in Fig.\ref{fig:res-rand-shakes}. The neural network performs well to even detect such interactions such as that might occur in an office corridor or a busy public place. The neural network was also tested for very rare cases by considering hand occlusion cases. Fig.\ref{fig:fakeshake} shows instances of such occlusions intentionally mimicking a handshake which the neural network avoids detecting. \par

\begin{figure*}[htb!]
    \centering
    \captionsetup[subfigure]{aboveskip=2pt, belowskip=2pt}
    
    \begin{subfigure}[t]{\columnwidth}
        \centering
        \includegraphics[width=\mywidth]{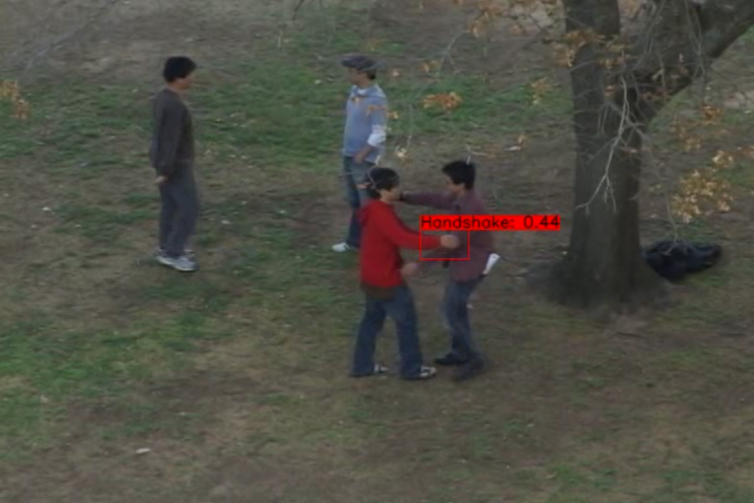}
        \caption{}
    \end{subfigure}%
    \hfill
     \begin{subfigure}[t]{\columnwidth}
        \centering
        \includegraphics[width=\mywidth]{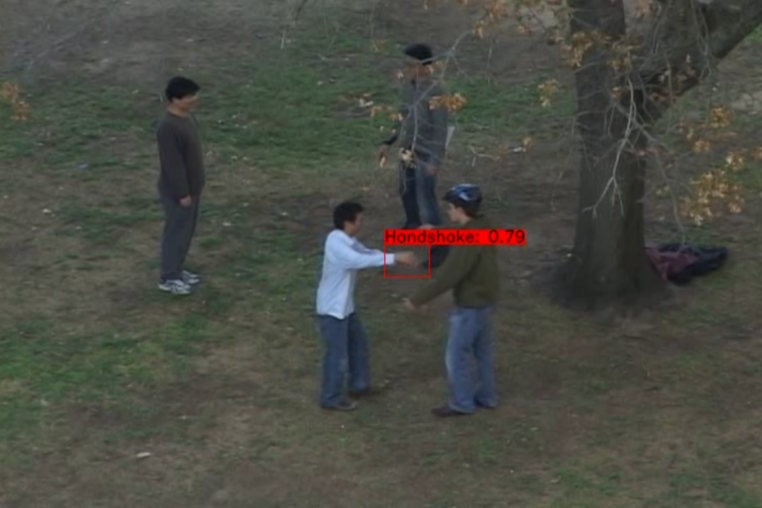}
        \caption{}
    \end{subfigure}%
    \hfill
     \begin{subfigure}[t]{\columnwidth}
        \centering
        \includegraphics[width=\mywidth]{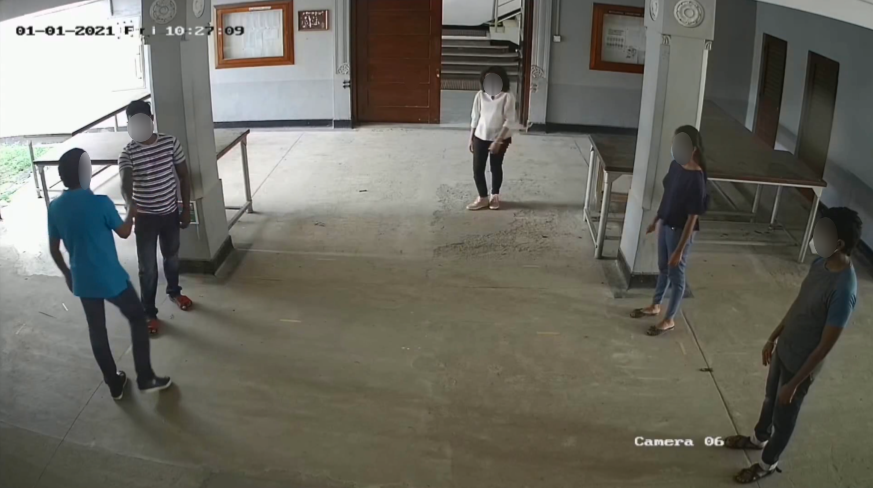}
        \caption{}
    \end{subfigure}%
    \hfill
     \begin{subfigure}[t]{\columnwidth}
        \centering
        \includegraphics[width=\mywidth]{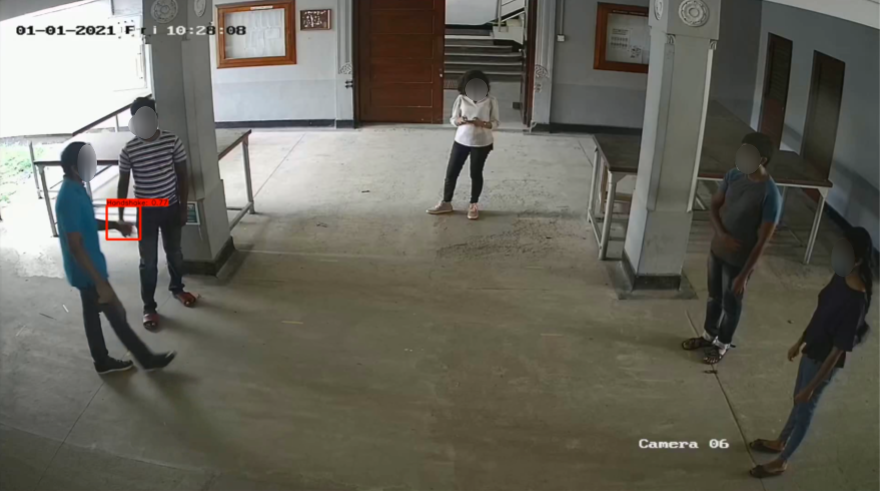}
        \caption{}
    \end{subfigure}%
    \hfill
    \begin{subfigure}[t]{\columnwidth}
        \centering
        \includegraphics[width=\mywidth]{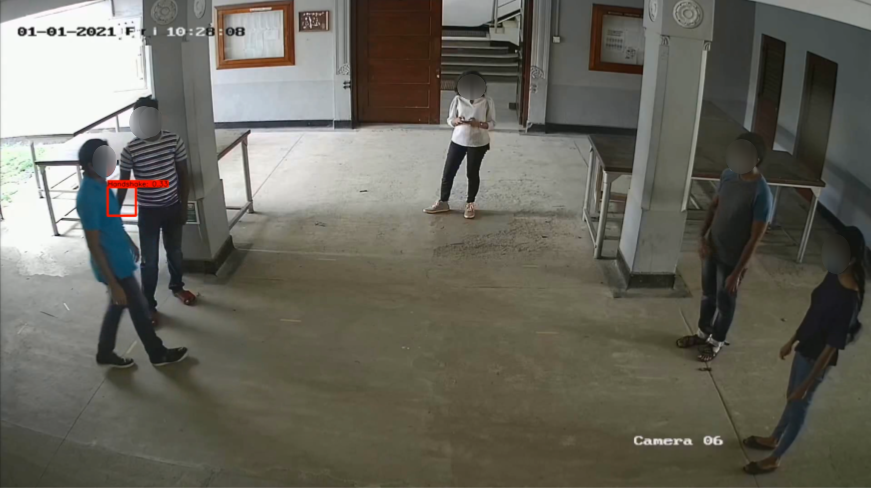}
        \caption{}
    \end{subfigure}%
    \hfill
     \begin{subfigure}[t]{\columnwidth}
        \centering
        \includegraphics[width=\mywidth]{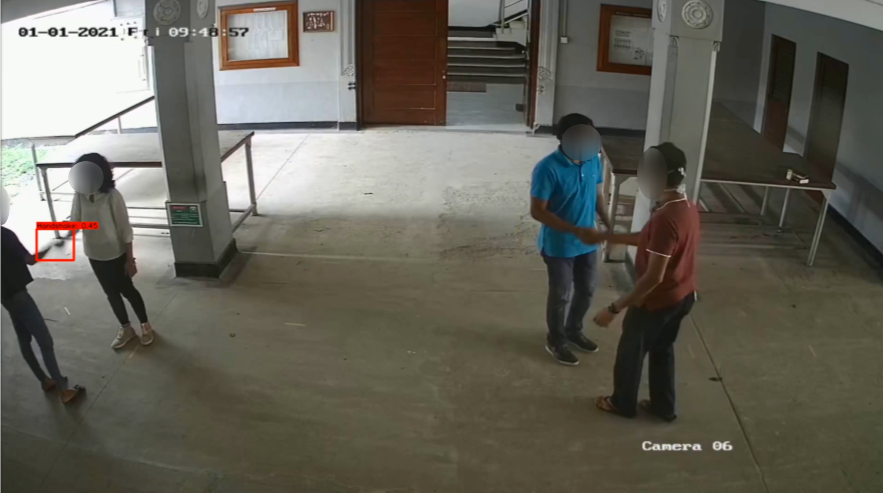}
        \caption{}
        \label{fig:deee-shakefail}
    \end{subfigure}
    \caption{Localization false positives and false negatives in the model.}
    \label{fig:shakefails}
\end{figure*}

Finally, the false positives of the neural network model in handshake interaction localization were analyzed. The false positives can be observed in Fig.\ref{fig:shakefails}. It can be observed that most errors occur during occlusion or in instances where the hand positions are similar to those during handshakes, ie: an outstretched hand and palm. Furthermore Fig.\ref{fig:deee-shakefail} depicts an instance where one handshake is identified whilst the other is not.

\section*{Conclusion}
Lack of human adherence to social distancing protocols is notably increasing thereby compounding the spread of \covid. This demands a scrutinized monitoring of human interactions in public spaces to identify and mitigate such violations of social distancing measures. In this paper, we present a neural network model to identify handshake interactions in realistic scenarios from CCTV footage in a multi-person setting. The neural network performance is validated by comparing its localization in 2 different datasets. The ability to detect and localize interactions in real-world settings and the detection of multiple interactions in a single frame affirm the robust nature of the model. The deployment of this model will enable us to identify social distancing violations in real-time and thereby create a framework to reduce such violations and mitigate the adverse impacts of \covid. As a deployable system, the model could be further improved to localize more challenging interactions such as hugs and kisses to combat the pandemic. 

\section*{Acknowledgements}
This work was funded by Lewis Power, Singapore and International Development Research Centre (IDRC), Canada. The authors are thankful to the volunteers for their support in creating the Shakes dataset.

\bibliographystyle{IEEEtran}

\end{document}